# RustSEG: Automated segmentation of corrosion using deep learning

B. Burton[1], W.T. Nash[2], N. Birbilis[1]

[1]College of Engineering and Computer Science, Australian National University, Acton. ACT. 2601. Australia
[2]WSP Canada Inc., 1935 Bollinger Road, Nanaimo, BC. V9S 5W9. Canada

**Abstract**

The inspection of infrastructure for corrosion remains a task that is typically performed manually by qualified engineers or inspectors. This task of inspection is laborious, slow, and often requires complex access. Recently, deep learning based algorithms have revealed promise and performance in the automatic detection of corrosion. However, to date, research regarding the segmentation of images for automated corrosion detection has been limited; due to the lack of availability of per-pixel labelled data sets which are required for model training. Herein, a novel deep learning approach (termed RustSEG) is presented, that can accurately segment images for automated corrosion detection – without the requirement of per-pixel labelled data sets for training. The RustSEG method will first, using deep learning techniques, determine if corrosion is present in an image (i.e. a classification task), and then if corrosion is present, the model will examine what pixels in the original image contributed to that classification decision. Finally, the method can refine its predictions into a pixel-level segmentation mask. In ideal cases, the method is able to generate precise masks of corrosion in images, demonstrating that the automated segmentation of corrosion without per-pixel training data is possible – addressing a significant hurdle in automated infrastructure inspection.

**Nomenclature**

*Abbreviations*
CAM        Class Activation Map
CNN        Convolutional Neural Network
CRF        Conditional Random Field
FCN        Fully convolutional network
GDP        Gross Domestic Product
NSFW       Not Safe for Work
ReLU       Rectified Linear Unit

**Introduction**

Corrosion of steel remains an ever-present concern for infrastructure and the built environment. If the corrosion of structures is left undetected, corrosion can propagate and lead to financial losses or the premature end-of-life for a structure. At present, the cost of corrosion remains significant. In 2016 NACE International put the total worldwide cost of corrosion at $2.5 trillion (USD) or 3.4% of global GDP [1][2]. In Australia, estimates by the Australasian Corrosion Association put the cost of corrosion in the tens of billions of dollars annually [3]. Corrosion inspections are typically carried out by qualified personnel, who inspect a structure for the presence of corrosion, and then, may recommend what action - if any - may be taken following corrosion detection. Such inspections are laborious, time-consuming and can sometimes require inspectors to access to hard-to-reach places.

The prospect of using computer vision methods applied to images (or video) of infrastructure for automated corrosion inspection, has recently been explored as an alternative to manual (human) inspection. Even in the most primitive application of automated corrosion inspection, which would be for the external and 'line of sight' view of infrastructure, computer vision can provide significant cost savings and removes the need for cumbersome access. A case in point, is that computer vision methods may be applied to video footage from a drone that can survey a building (including at significant height) or survey at the whole-of-plant scale.

Physical inspections can now be performed by unmanned aerial vehicles (UAVs) and other remote cameras. At present, such footage is then reviewed 'manually' by qualified engineers [4, 5]. Depending on the size of the asset this could be hundreds of hours of footage. This, combined with the subjectivity of human review, makes this a ripe area for automation. For such tasks, researchers have explored high performance image-based corrosion detection algorithms, built using deep-learning models. To date, no 'human level performance' generic corrosion segmentation algorithm has been developed (or at least, publicly available). One major reason is the lack of large accurate corrosion segmentation data sets, that are required for the model training. Furthermore, training data used in discrete studies reported to date, is usually highly specific to the environment and context of which that data was collected – as opposed to general. The purpose of the work herein is to prepare a general deep-learning model that can perform in a range of context/environments, and also to answer the key research question, which is: C*an a deep learning based corrosion segmentation model – that does not require segmented training data - achieve accurate results and repeatability for automated corrosion segmentation?*

Early related research in computer vision (CV) for infrastructure inspection utilised colour information to pursue a consistent method for determining the amount of corrosion on bridges [6]. Those CV methods required uniform backgrounds (i.e., painted metal) and were easily confused by artifacts in the images that may have included shadows, or more nuanced defects like flaking paint or rust stains. Such work was expanded to include texture information when the background was of a similar colour to corrosion (i.e. bridges painted red) [7]. Many of the early segmentation methods focused on situations where all images inspected were known to have corrosion present and only the segmentation masks were being assessed. *Shen et al*. [7] outlined a method of initially classifying corrosion in an image then based on this result, trying to generate segmentation masks of this corrosion. Such methods required sanitised inputs and often were only accurate on images of uniform background.

With advancements in deep learning for tasks like classification [8] and segmentation for non-task specific operations [9], researchers also focused on deep learning models to identify corrosion from images. In 2016, Petricca and co-workers [10] showed that deep learning models outperform traditional CV techniques in the classification of corrosion. Deep learning models have been able to demonstrate 95% accuracy for task specific corrosion identification such as corrosion related defects on railways [11]. Other researchers also demonstrated feasible corrosion inspections using non-convolution based deep learning methods in combination with drone footage, could be applied to ballast tanks in ships [12]. Liu et. al. [13] were able to adapt VGG-16 [14], an 'off-the-shelf' successful method of generic segmentation, to the specific segmentation of corrosion in ballast tanks in ships with over 80% accuracy. Deep learning has also been applied with different levels of success to non-imaged based corrosion detection methods such as utilising Eddy currents in metal plates [18], thermal responses in aluminium [19], and electrochemical data in steel [20]. Researchers have also applied deep learning models to other non-metallic defect detection problems such as concrete cracking [15].

Recently, much like with the general field of CV, deep learning researchers have looked to convolutional neural networks (CNNs) as the leading method for imaged based deep learning methods [15, 16]. In classification, researchers have been able to achieve +90% accuracy for generic corrosion detection using only deep learning methods [15, 17]. However, as noted, the segmentation of corrosion has proven more challenging for deep learning based methods, due to the unavailability of suitable training data [15]. Deep learning based methods for corrosion require large, accurate, labelled data sets. In the area of corrosion detection and segmentation such data sets do not exist [15]. This problem is clear when comparing segmentation of common objects to the segmentation of corrosion. The popular data set MS COCO [21] contains very accurately labelled common objects. The MS COCO dataset contains 328,000 images and 2,500,000 instances accurately labelled. This data set was trained using paid crowdsourced workers from the *Amazon Mechanical Turk* service. Each image was segmented by eight different people to ensure accuracy. This labelling processes was simplified because the MS COCO data set only contains objects so common that they could be identified by a four year old [21]. In contrast the work reported in [5] used only 608 images with 4,631 instances of corrosion, hand labelled by the researchers themselves.

An analysis by researchers in [22] suggested that an average of 65,000 labelled images would be required to make a human comparable generic corrosion segmentation method using deep learning segmentation models such as VGG-16 [14]. In comparison, to the authors knowledge, the largest per-pixel labelled corrosion data set reported in literature was described by Fondevik and co-workers [5], and contained 608 images, however unlike MS COCO, such data is not publicly available. The researchers in [13] used a larger, task specific, corrosion segmentation data set (1900 images), but again this was not made available to the public. Fondevik and co-workers [5] highlighted some issues with creating segmented corrosion data sets. Certain characteristics of corrosion make it extremely difficult to mask accurately, even by experts. Non-uniform edges, rust stains, flaking paint and shadows in combination with differences of opinions by labellers, made the exact determinations of boundaries essentially impossible. However, Nash and co-workers [23] revealed that larger, less accurate, data sets perform better for training in corrosion detection than small expertly labelled sets. The same researchers showed in [22] that this also applies to segmentation. Additionally, Nash and co-workers [17] also developed an accurate deep learning based corrosion 'detector' (i.e. classification tool) that used crowdsourced data – and with no information on the level of expertise of the users. However, to the present authors knowledge to date, no methods have been developed that uses crowdsourced per-pixel corrosion data for training.

The task of corrosion segmentation has been of interest for some time – with reports stating that as early as the late 1990's researchers were using image-based corrosion segmentation methods for defect detection in painted bridges [7]. Non-deep learning method such as those described by Ling et al. [24] have been successful at generating segmentation masks to determine the amount of rust present in an image from inspection photos of painted steel. However, as deep learning methods have evolved, researchers have developed a range of models to address the issue of segmentation. Recently, several groups have presented deep-learning models for corrosion segmentation to a range with varying levels of performance [5, 13, 22]. All these methods used small highly accurate labelled data sets and combine them with transfer learning. These data sets require expert notation and are extremely time consuming. In a study by Nash and co-workers [23] it was noted that a larger volume of data (including 'noisy' data) could perform better than a smaller data set with only a few highly accurate data samples - in the context of generic corrosion segmentation. It was posited in [22] that with 65,000 images a human level model could be created. It was also claimed with combination of two data sets, one so called 'perfect' and one imperfect could reduce this significantly. In a similar context, [25] and [26] have developed an accurate model for generating bounding boxes for specialised defect detection tasks. The method described in [25] combined deep learning and traditional computer vision methods to generate a bounding box approach.

To date, researchers have demonstrated success in visualising the output of so-called convolutional neural networks (CNNs). Yosinski et al. [27] were able to produce localisation clues on a class from the kernels of the convolutional layers. A discriminative localisation approach, known as Class Activation Mapping (CAM) and presented by Zhou and co-workers [28], demonstrated that using the last convolutional layer, the predicted class and hidden layers – it is possible to visualise the pixels that have greatest effect on the prediction in a multi class classifier. An approach termed Gradient-weighted Class Activation Mapping (Grad-CAM) [29] expanded the work of Zhou [28] - to produce highly localised heatmaps of the pixels that have the greatest effect on predicting classes, whilst removing the need for any retraining. Thereafter, Grad-CAM++ [30] was developed to overcome some of the shortfalls from the original methodology when applied to images that contain multiple instances of the target class. Most recently, in 2021, researchers used Grad-CAM to demystify neural network detections of defects in airplane fuselages to guide human technicians in repairs [31]. The ability to accurately segment images *without* the use of segmented training data remains of ongoing interest to researchers [32], on the basis that it decreases the most complex aspect of model training (including vast benefits in time, effort, and cost). The fully convolution network (FCN) presented in 2014 by Long and co-workers [9] was the first fully convolutional end-to-end network trained for semantic segmentation. It adapted then state of the art classifier networks to produce heat maps that were then back propagated with annotated training data for accurate segment results. So called "U-Net" [33] further developed upon the FCN an and achieved state of the art results on medical segmentation on small training datasets (~30 images). However, both such methods (FCN and U-Net) still required some segmented training data. Other researchers [34, 35] have been very successful at producing segmentation from bounding boxes and from weak localisation clues (partial segmentation) [36, 37]. More recently the seed, expand, an constrain (SEC) approach presented by Kolesnikov and C. H. Lampert [38] demonstrated a method that uses a weak localisation clue with as information from a classifier network to determine how large a mask will likely be; then developed a method for "growing" this mask to the boundary of the object. In [29] they demonstrated that the Grad-CAM method could be used as the seed in the SEC method to produce highly accurate segmentations on the popular PASCAL VOC 2012 [39] data set.

**Methodology**

Herein, the RustSEG method uses a novel approach to combine multiple known computer vision concepts, with customisation to create an end-to-end approach for segmentation of corrosion in images without the requirement for any per-pixel training data. For the purposes of this algorithm, corrosion was limited to ferrous corrosion, commonly known as 'rust'.

*Data Set*

At present, no large-scale publicly available corrosion data sets exist. To construct a large training data set for the work herein, publicly available photos from the popular photo sharing website *flickr.com* were used (much like the PASCAL VOC 2012 [39] dataset). This data set was constructed using keyword searches for corrosion including: "corrosion", "rust", "rusty car", "rusty bolt". For images not containing corrosion (i.e., 'random' images) keyword searches were chosen using two methods. Firstly, categories known to trick or confuse both humans and CNNs (as discussed in [5, 17]), as to the presence of corrosion, were sought - including terms like: "bricks", "wooden boats", "graffiti" and "brioche". The other type of not-corrosion images representing everyday objects and scenes, the search terms for this include: "people", "outside", "homes" and "animals".

The data sets were lightly checked mostly to remove animated or rendered images, NSFW images and images of the music band also called "corrosion". The final size of the data set for training was 7000 "corrosion" images and 9000 "not-corrosion" images. With an additional 600 corrosion and 600 not-corrosion for testing. An additional 1200 images were obtained from the training dataset of corrosiondetector.com [17] and other academic sources. These were set aside as an expertly labelled test data set. It is acknowledged that false positives and false negatives almost certainly exist within the data set, however following the work described in [23], it is assumed that larger data sets are better than smaller highly accurate ones.

It should be noted that *flickr.com* is popular with professional photographers. As a result, many images have been manipulated (i.e., saturations changed, filters applied, colours manipulated) and a small number of images contain water marks. The effects of this on RustSEG performance are discussed further below, and example images for the dataset used herein are shown in Figure 1.

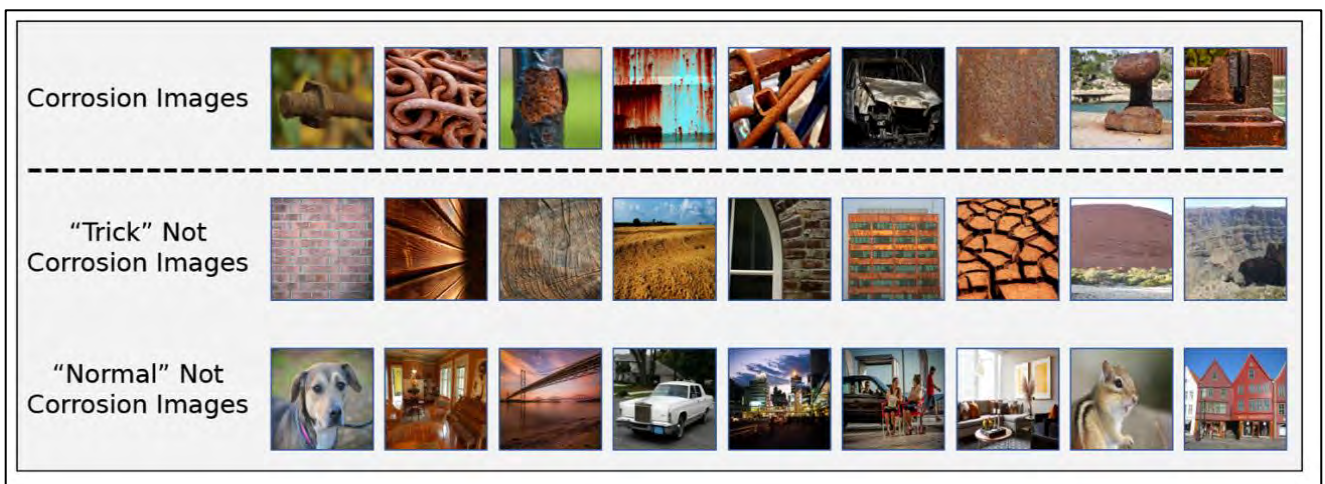

*Figure 1.* Examples from training data set including: (top) images of corrosion, (middle) "trick" images that are not corrosion, but may be misinterpreted as of similar appearance to corrosion, (bottom) not-corrosion images

### Workflow (Classify, Localise, Refine)

To produce segmentation masks using 'no per pixel' training data a simple three step workflow was constructed: classify, localise and refine. A general description of the workflow is provided, with each of the three steps then elaborated.

Firstly, input images are rescaled to 572×572 pixels and are fed into a CNN-based binary classifier. If corrosion is detected with a greater than 50% corrosion prediction the images are passed to the localisation function. This localisation is an adaptation of the Grad-CAM++ method and returns a 28×28 pixel "heatmap" of corrosion in the image. The heatmap is then up sampled using standard linear interpolation to the original 572×572 resolution. A threshold filter (described in Eqns. 8 and 9) is applied to remove 'low scoring' pixels. This mask is then modified (refined) with a conditional random field (CRF) for 25 epochs. Finally, the mask is superimposed on the original image and returned to the user.

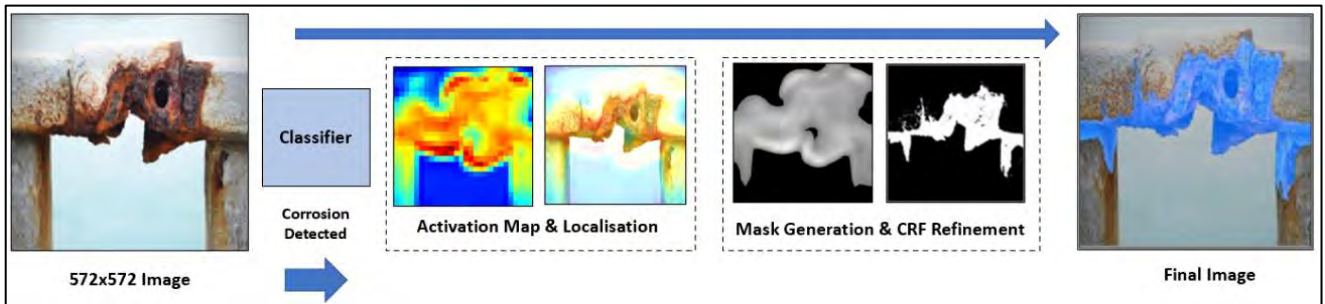

*Figure 2. Schematic of the end-to-end workflow of the RustSEG method*

### Classify: CNN- Classifier

To construct the classifier for RustSEG, the methodology described in [17] was used, which is based on the encoder side of U-Net described in [33]. The classifier model was constructed using Python, utilising the *TensorFlow* API. It consists of ten layers in five convolution sections with each convolution layer being followed by a rectified linear unit (seen in Eqn. 1) to bind the results as positive.

$$f(x) = \max(0, x) \qquad (1)$$

Each convolution layer utilises a 3×3 kernel and the 'valid' padding method (which assumes all dimensions/scaling in the images being processed are valid), therefore each layer decreases by one pixel on each side of the perimeter. Between every section there is a 2×2 max pool operation. In effect this shrinks the input height and width by half and doubles the depth. Finally, the final layer is flattened and fed into a one neuron layer with a sigmoid activation function (seen in Eqn. 2).

$$f(x) = \frac{1}{1+e^{-x}} \qquad (2)$$

A value between 0 and 1 is produced. 0 for 100% prediction of corrosion, 1 for 100% confidence of not-corrosion and 0.5 represents unsure. Having this range of confidence allows for a threshold of confidence to be set in the future for more critical applications.

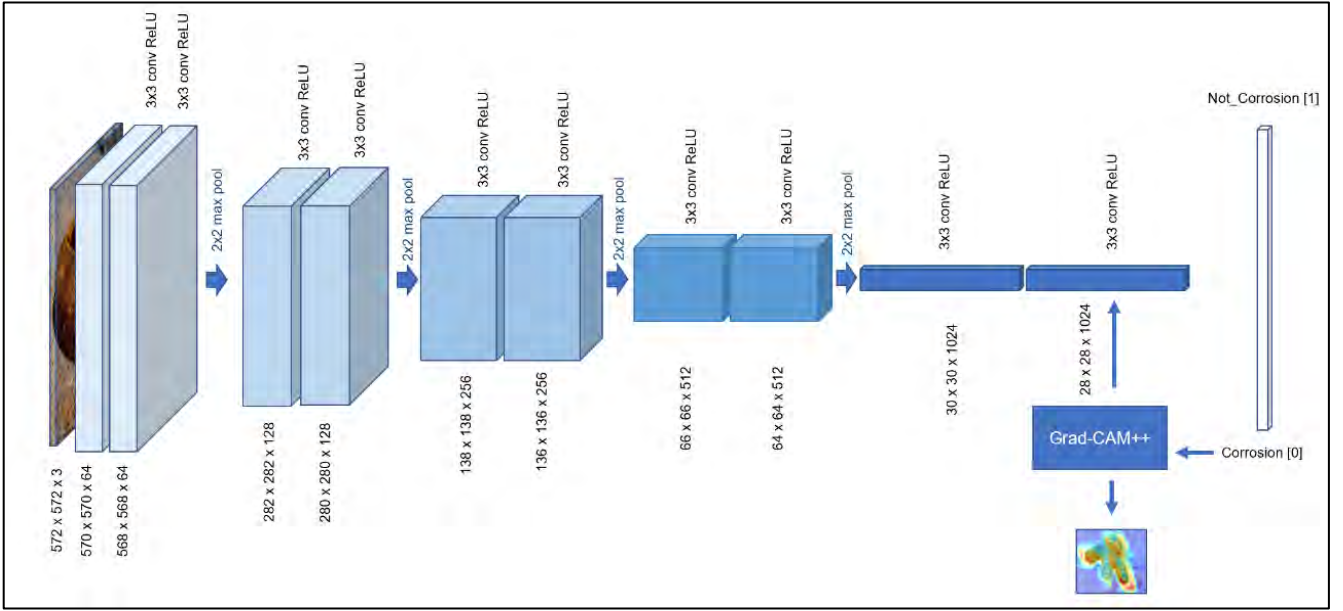

*Figure 3. Block diagram of the CNN used for the RustSEG method in this work*

The block diagram of the CNN used herein is shown in Figure 3. For training the RustSEG model a slightly different approach to that of Nash *et. al.* [17], was used. To avoid overfitting, image augmentation was implemented on the dataset, using random flipping and rotation (-0.2 to 0.2 rads). This slight data augmentation was deemed appropriate, as corrosion has no discernible orientation, so adjusting its relative position in the image should not affect the model's localisation ability. The image augmentation was performed outside of the model, before images were passed to the model – allowing integration with the Grad-CAM++ module to be simpler, and with no effect on the accuracy compared to the *TensorFlow* in-model data augmentation method. The approach herein does add a small increase in overhead for training via *TensorFlow* implementation. A balanced data set of 7000 corrosion images and 7000 non corrosion images was used. 15% of each class (1050:1050) was separated out as a validation set to monitor training for overfitting.

He-Kaiming uniform initialization was used for the model weights [40]. For training, a learning rate of 0.00001 was used along with the Adam optimizer [41] and binary cross entropy loss. Training was undertaken for 35 epochs.

### *Localise: Grad-CAM++ (Heat Map)*

If the classifier model detects corrosion, the associated image is passed to the Grad-CAM++ module (Grad-CAM++ described in [30]). It is noted the that the predecessor model, known as Grad-CAM [29], remains more widely used – even today - for generating rich activation maps. However, the Grad-CAM method has two key issues that negatively affect its ability to localise corrosion in images. Firstly, it performs poorly when multiple instances of the same class appear in a single image (which is common with the images used herein, and in the case of corrosion more broadly). Secondly, it often fails to highlight the full shape of a class, instead generating a partial highlight. Grad-CAM++ overcomes these two issues - generating more appropriate heatmaps for investigating corrosion.

To understand how the Grad-CAM++ method works, an understanding of the Grad-CAM predecessor is required [29]. The Grad-CAM method finds the gradient of the score of the winning (maximal) class (from the output neuron) with respect to the activation maps of the last convolutional layer (before the ReLU is applied). These gradients are global-average-pooled over the height and width of each convolutional layer. Mathematically this can be represented as:

$$\boldsymbol{A}^k \in \mathbb{R}^{u \times v} \qquad (3)$$

Where $A^k$ is the $k^{th}$ feature map of the last convolutional layer ($k = 1, ..., K$). $u$ and $v$ represent the height and width of the same feature maps. In the case of the model used in this approach $k = (1, ..., 1024)$ and $u, v = 28$. Thus, using this Grad-CAM can be considered as:

$$L_{Grad-CAM} = ReLU\left(\sum_{k=1}^{K} \alpha_k^c A^k\right) \quad (4)$$

Where the ReLU is the same function described in Eqn. 1, and $\alpha_k^c$ is the global average pooling of the gradients of the output class and the feature maps as described below in Eqn 5. The ReLU function is applied as shown in [29] and in [30] including gradients that have a positive influence on predictions resulting in better results than the combination of negative ones. This is intuitive, because the number of features that make up 'corrosion' although possibly large - is finite. However, the number of features that makes up the class 'not-corrosion' is effectively infinite.

$$\alpha_k^c = \underbrace{\frac{1}{u \times v} \sum_i \sum_j}_{Global\ average\ pooling} \frac{\partial y^c}{\partial A_{i,j}^k} \quad (5)$$

Where $\frac{\partial y^c}{\partial A_{i,j}^k}$ is the gradient of the score for class $c$, $y^c$ (for RustSEG, this is always the class 'corrosion') with respect to the $i^{th}, j^{th}$ equivalent pixel position of the $k^{th}$ feature map.

Eqn 5 is where Grad-CAM++ has changes that make the method more robust. It was shown [30] that by taking an unweighted global average pool of the partial derivative, it is possible to 'wash out' multiple instances of the same class including parts of the same instances - leaving only partial localisations of instances. Grad-CAM++ also proposes excluding negative gradients before the combination with Eqn 4 and shows this in an empirical analysis in their paper [30]. Instead of a global average pooling, Grad-CAM++ proposes a weighted average pool to replace Eqn 5 as:

$$\alpha_k^c = w_{i,j}^{k,c} \sum_i \sum_j ReLU\left(\frac{\partial y^c}{\partial A_{i,j}^k}\right) \quad (6)$$

Where $w_{i,j}^{k,c}$ as derived in [30] is:

$$w_{i,j}^{k,c} = \frac{\frac{\partial^2 y^c}{\left(\partial A_{i,j}^k\right)^2}}{2\frac{\partial^2 y^c}{\left(\partial A_{i,j}^k\right)^2} + \sum_i \sum_j A_{i,j}^k \left(\frac{\partial^3 y^c}{\left(\partial A_{i,j}^k\right)^3}\right)} \quad (7)$$

The full derivation for the weights shown in Eqn 7 were not strictly relevant for the implementation in RustSEG with the exception that Grad-CAM++ also allows the calculation of the second and third order gradients for each heatmap generated. Such extra order terms that would come at significant computational expense – are considered to be explored in the future. For the benefit of readers, it is noted that when implemented in *TensorFlow* the RustSEG tool uses three nested `tf.GradientTape()` loops passing the calculated gradients from the inner most loop out for higher order derivatives needed.

### *Refine: Post-Processing (mask refinement)*

For this project, the post-processing methods that were developed and employed, are described separately below.

## Rescale and Filter

The Grad-CAM++ method returns a 28×28 heatmap of values (floats) between 0 and 1. This is then resampled using linear interpolation to the original 572×572 image size. The heatmaps will generally not return a 'perfect' 0 for no corrosion, or a 1 for corrosion. To therefore 'clean' the mask and remove small scores, a dynamic threshold filter is applied. The threshold is calculated according to Eqn. 8.

$$T = 1 - (max(H) - \bar{H}) \qquad (8)$$

Where $H$ is the 28×28 heatmap, $max(H)$ is the maximum value of the heatmap, $\bar{H}$ is the mean value of the heatmap and $T$ is the threshold for the filter. The filter is applied according to Eqn. 9,

$$M_{i,j} = \begin{cases} H_{i,j} & if \quad H_{i,j} \geq T \\ 0 & if \quad H_{i,j} < T \end{cases} \qquad (9)$$

Where $M_{i,j}$ is the new filtered mask. The mask is then scaled from being between 0 and 1 to being between 0 and 255. It is then converted out of a 32-bit precision float (float32) data type to an unsigned 8-bit integer (uint8). The conversion is done using nearest rounding. Any information lost from this step was deemed trivial. An example of the effect and operation of threshold filters is given in Figure 4.

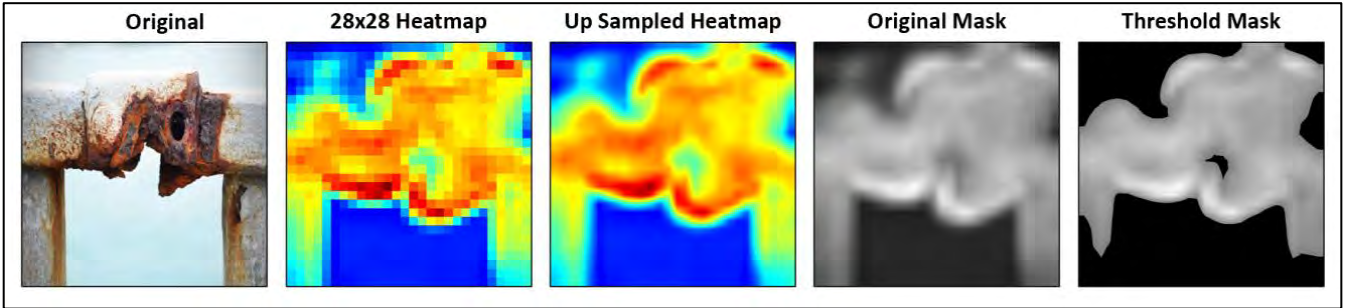

***Figure 4.*** *Schematic revealing the effect of dynamic threshold filters on mask*

## Pure CRF

A conditional random field (CRF) can be utilised to improve segmentation masks. A CRF takes into account colour, texture and contiguity of an input seed, and will grow or shrink the region based on bordering pixels and hyper parameters of these three criteria. The version used herein is based on *Krähenbühl and Koltun's* method described in [42], and was implemented through a publicly available module called PyDenseCRF [43]. The derivation of how the CRF works can be found in [42], and given by:

$$CRF_n^{x,y} = \underbrace{(Appearance\ Kernel)}_{\sigma_A^{x,y}, \sigma_A^{rgb}, \mu_A} + \underbrace{(Smoothness\ Kernel)}_{\sigma_S^{x,y}, \mu_S} \qquad (10)$$

Where $CRF_n^{x,y}$ is the CRF's determination at the $n^{th}$ epoch if the pixel at $(x, y)$ is to be included in the mask. It archives this by passing two filters over the image, a simple gaussian filter and a bilateral filter. Together the filters have 5 tuneable parameters: $\sigma_A^{x,y}, \sigma_A^{rgb}, \mu_A, \sigma_S^{x,y}, \mu_S$. Where $\sigma_A^{x,y}$ is the standard deviation in $x$ and $y$ directions (how large the filter is), $\sigma_A^{rgb}$ is the standard deviation in RGB channels and $\mu_A$ is the Potts compatibility term that penalises masks with non-contiguous shapes within the kernel of the filter. $\sigma_S^{x,y}$ is the standard deviation on the gaussian kernel size and the $\mu_S$ term is the compatibility term but can be thought of as the strength of the guassian filter. For each pixel along the edges of the current mask, the filtered values are compared to the values inside the mask. In the CRF these are called energies. If the energy is high enough outside the mask it expands or if they are too low the mask shrinks.

A CRF can be implemented in one of two ways for image segmentation. They can be trained where the hyperparameters are set dynamically using training ground truth data - often incorporated into the training of the neural network they are partnered with. Alternatively, they can be tuned based on their effect. Due to the lack of segmentation training data, herein the hyperparameters were set through an extensive trial and error process. An example of CRF mask refinement is provided in Figure 5.

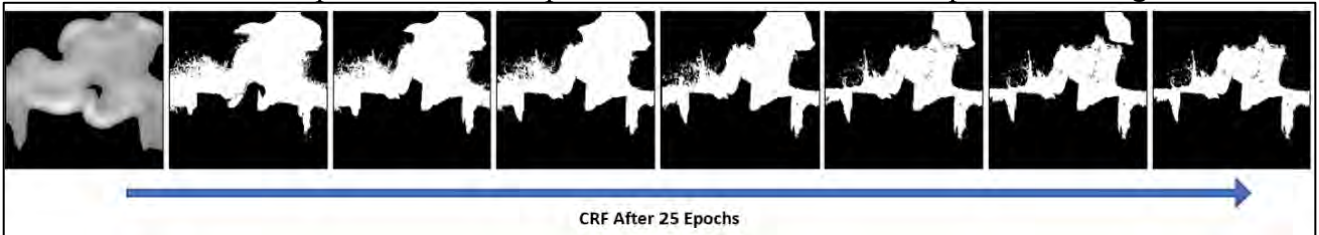

*Figure 5. Example of CRF mask refinement, occurring over 25 epochs*

### K-Means + CRF (Advanced)

When tuning the aforementioned CRF, in order to achieve excellent results for automated corrosion segmentation, it was noted that a heavier weighting was placed on the colour of pixels rather than their position. This reflects that rust can appear anywhere in an image but generally exists within a finite colour range. However, if the initial mask generated by the heatmap overestimates the shape of the corrosion and collects some of the background colour, the CRF will include this information to expand the region. To combat this phenomenon, a more aggressive initial mask refinement was proposed. Using the same initial filtering described above, then using the K-means algorithm generated with random centres, the image is segmented into two regions. The filtered mask is then compared to the two segmented regions. The K-means mask that covers the most (by area) of the heatmap mask is selected. This is then used to further 'trim' the heat map mask. A simple 5×5 diamond erosion is then applied to remove any spurious inclusions - after which the mask is passed to the CRF module (however only for 10 epochs to reduce computation burden required). This approach is called the 'Advanced' method for the utilisation of RustSEG, and is depicted in Figure 6.

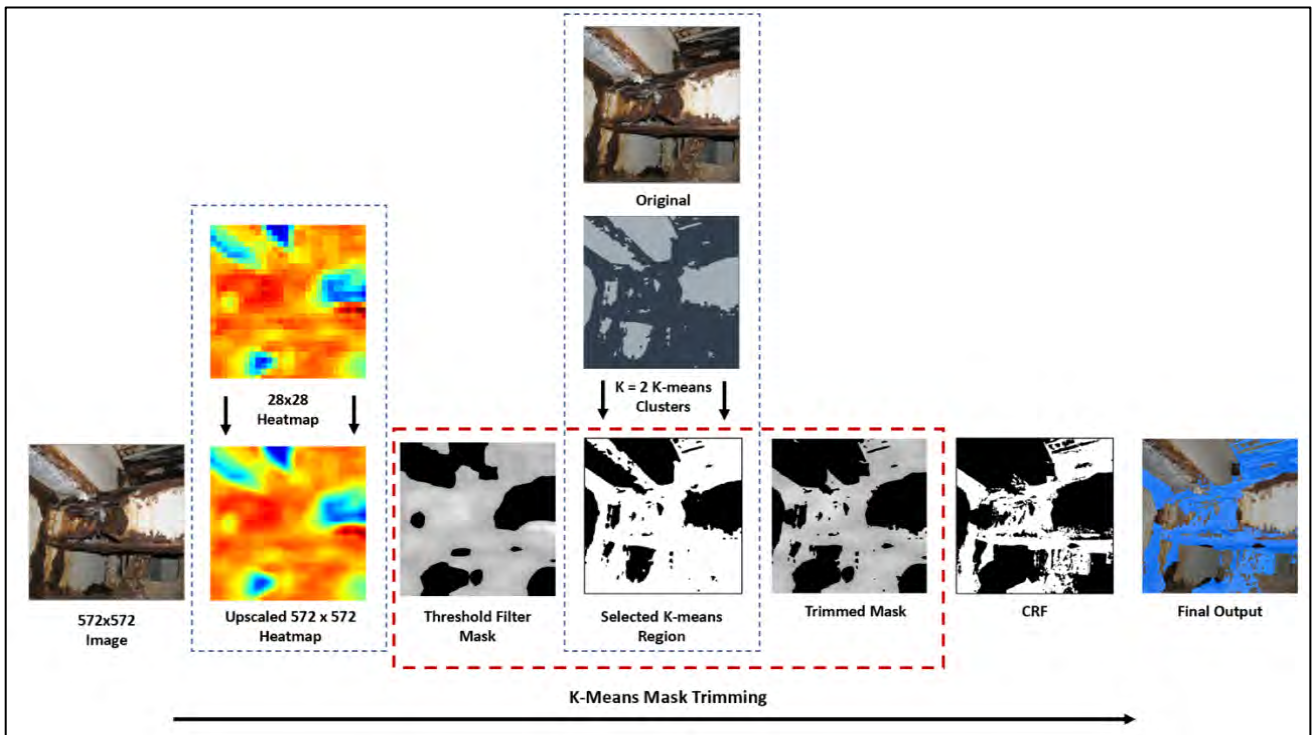

*Figure 6.* End-to-end work flow schematic of the RustSEG 'Advanced' method that utilises K-means clustering for the rapid trimming of masks

**Results**

The RustSEG method as tested by the authors was found to be capable of generating per pixel segmentation masks of ferrous corrosion, from unseen images – and without any per pixel training data. However, it is noted that on occasion, the performance of the RustSEG masks could be impacted by shadows and other image defects. Owing to the lack of openly available data sets of corrosion segmentation, the analysis of results herein includes qualitative descriptions. The results are presented according to component of the RustSEG workflow (Fig. 2).

### *Classifier Results*

The CNN model used for this project was similar to the method described in [17], which was utilised for the 'corrosiondetector' tool. Evaluation of the classifier utilised herein, was done via the utilisation of two test data sets. The first test set was 1600 images with a 1:1 split of corrosion to not corrosion, obtained from the authors and the original training set of [17] – and for the purpose of evaluation, they should be considered expertly labelled. The second test set was 1000 images with a 1:1 class split obtained using the same method as the aforementioned training dataset (obtained from *Flickr*), with care taken to make ensure the test set contained none of the same images as the training or validation data sets.

For determining the success of the classifier five common accuracy parameters were calculated. Accuracy, area under the ROC curve, precision, recall and F1.

Accuracy was defined as the number of correct predictions over the total number of predictions.

$$Accuracy = \frac{TP+TN}{TP+FP+TN+FN} \qquad (11)$$

Where $TP$ = true positive, $TN$ = true negative, $FP$ = false positive and $FN$ = false negative.

Area under the ROC curve can be thought of as how much better the model is than random chance. In a binary classifier 1 is considered a perfect model and 0.5 is equivalent to "flipping a coin".

Precision is defined as:

$$Precision = \frac{TP}{TP+FP} \qquad (12)$$

Precision can be considered as "measure of how often the model suggests something is corrosion, it actually is".

Recall is calculated as:

$$Recall = \frac{TP}{TP+FN} \qquad (13)$$

Recall can be considered as "for every actual image of corrosion given to model, how often it predicts that the image is corrosion".

The classification F1 score is a combination of precision and recall in the form of their harmonic mean:

$$F1 = 2 \times \frac{Precision \times Recall}{Precision+Recall} \qquad (14)$$

As precision and recall often come at the cost of the other, maximising F1 is a good indicator for robust classification.

The above parameters allow for more nuanced understanding of the classification model accuracy, in particular for assessing any imbalanced data. The results obtained herein for the classifier are presented in Table 1, for the two different test data sets explored. For each training exercise shown below, training took approximately 7 hours for 35 epochs on an Nvidia Tesla P-100 16GB (PCIe).

*Table 1.* CNN classifier results (Note: these results only show the results for classification, not segmentation)

| Test Data Set Used | Parameter | Score |
|---|---|---|
| Expertly labelled test data set (1200 images) | Accuracy | 77.83% |
| | Area Under Curve | 86.17% |
| | Precision | 81.27 % |
| | Recall | 72.33% |
| | F1 | 76.54 % |
| *Flickr* test data set (1000 images) | Accuracy | 86.81% |
| | Area Under Curve | 94.82 % |
| | Precision | 89.38 % |
| | Recall | 83.86 % |
| | F1 | 86.53 % |

### *Heat Map Results*

There is no standard method for assessing the accuracy of class based heatmaps. For this analysis they were assessed qualitatively. The same supplied test set used for assessment of the classifier was used for assessing the performance of RustSEGs implementation of Grad-CAM++ in localising corrosion in images. As e seen in

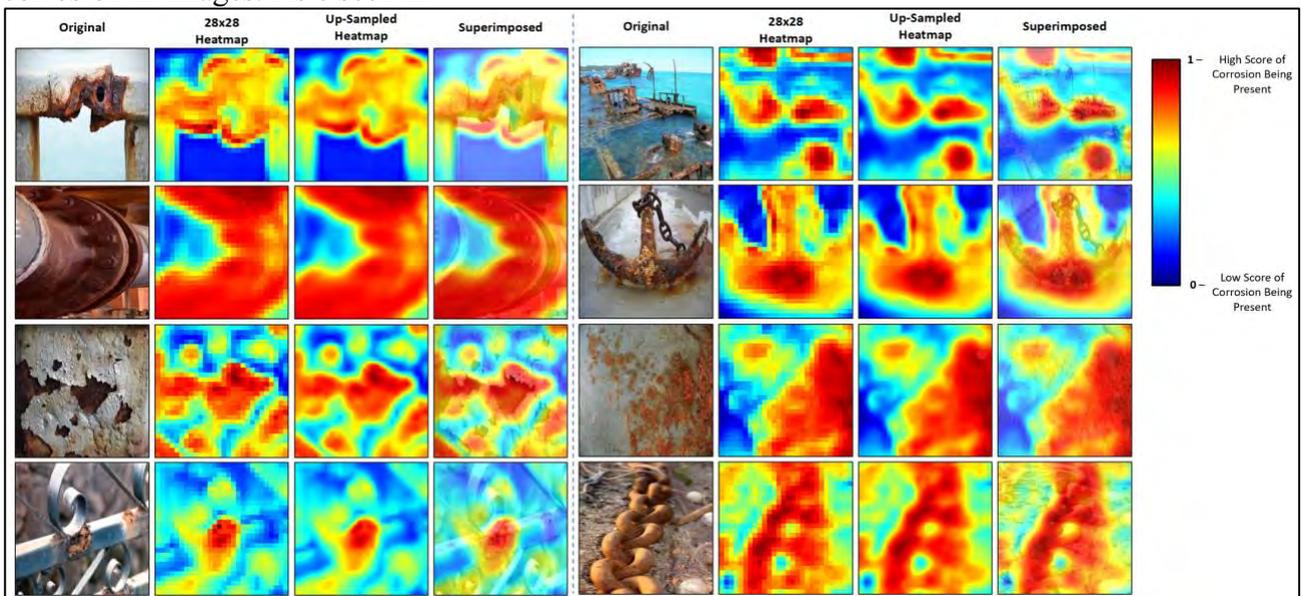

*Figure 7*, in some settings this method is very effective at locating the corrosion in the image. The method appears to perform very well in images that have blurred backgrounds or rust emerging through painted surfaces.

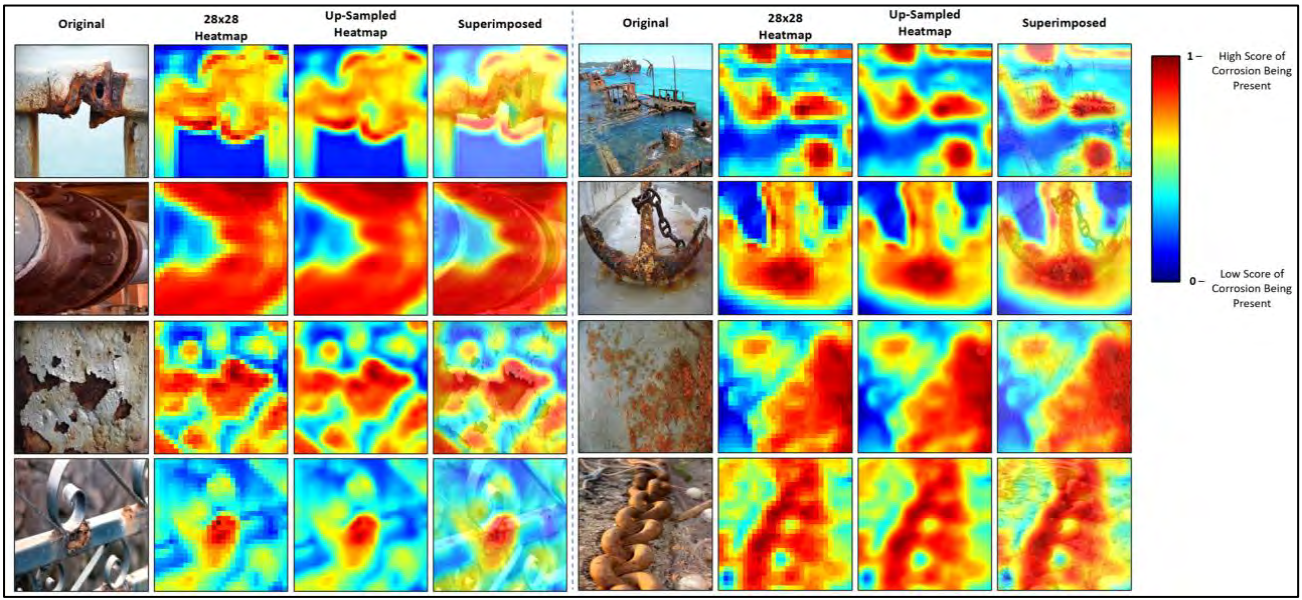

*Figure 7. Highly accurate heatmaps generated for localisation*

Images that contain harsh shadows bordering the areas of corrosion appear to frustrate the method. In some cases, the method is able to also accurately locate the corrosion, but can be mistaken and also highlights the adjacent shadow. In some other cases the method appears to preferentially highlight the shadow – as demonstrated (Figure 8).

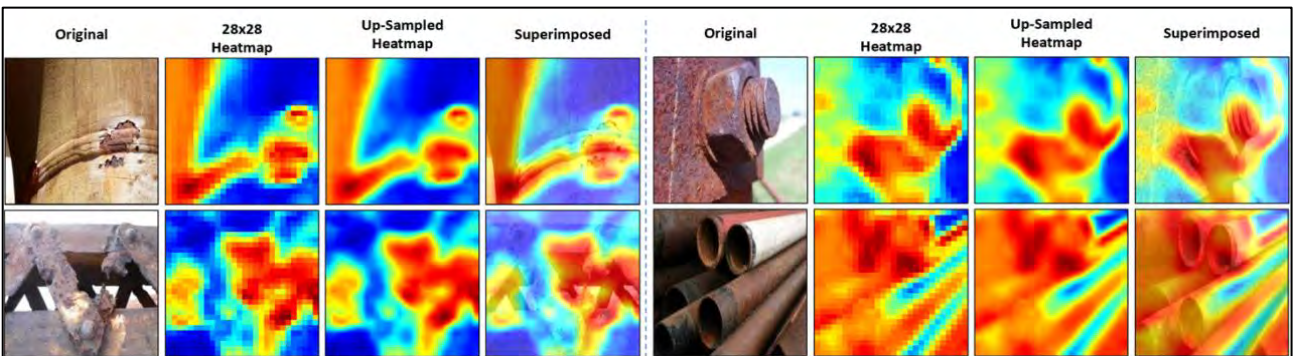

*Figure 8. Heat maps that show poor performance due to shadows*

A common case that was discovered to cause the RustSEG model frustration was when the entire input image being analysed contains corrosion. It would appear that the methodology to be successfully deployed, some context is required in the image to permit the determination of corrosion location. As revealed in Figure 9, it appears as though the heatmap does actually generate the correct output (i.e. corrosion highlighted). It is noted that the right side of Figure 9 also reveals images that only contain corrosion - but are not necessarily only a 'close up' of corrosion – revealing frustration in RustSEG performance. The performance issues in Figure 9 (and other like examples not shown) appear to be a failure of the localisation process, as the CNN was able to correctly classify all such examples (including those in Figure 9) as corrosion.

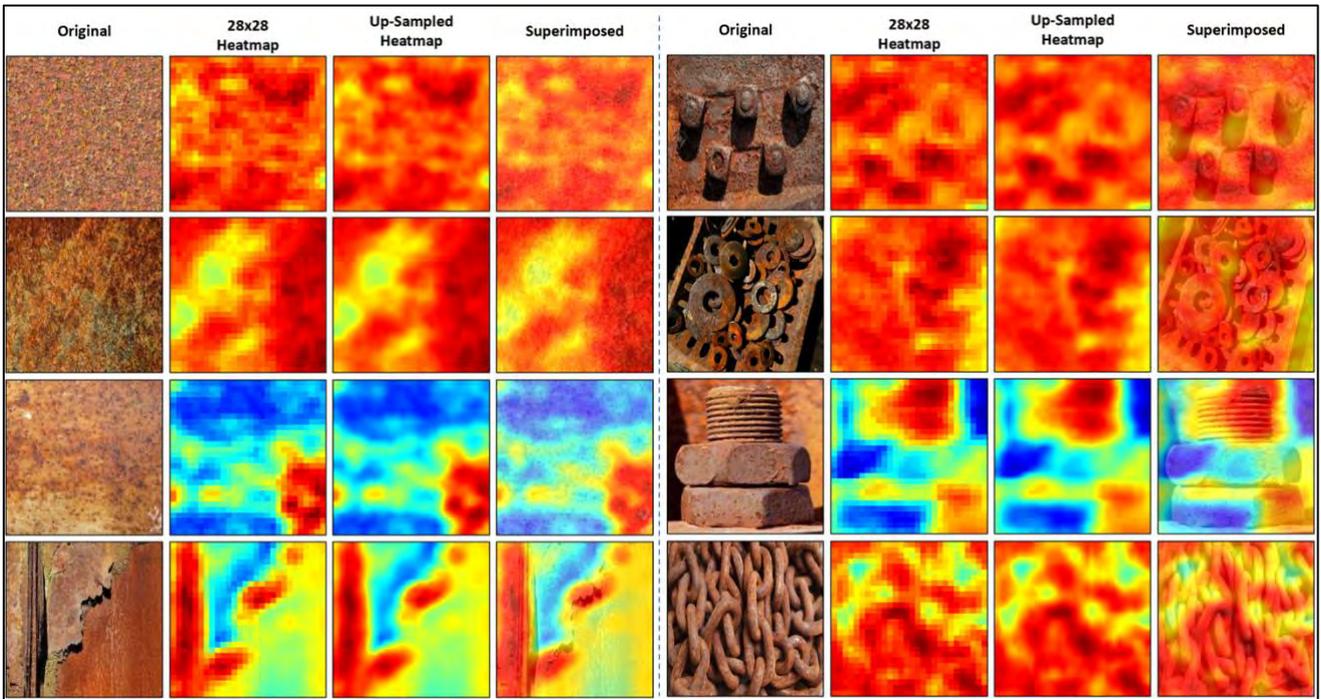
*Figure 9. Heatmap that show inconsistent results on image that only contain corrosion.*

The localisation, much like the classification of corrosion, whilst consistently well-performing, includes example that are not always perfect. Figure 10 reveals a random selection of test images that the system correctly identified as corrosion, but failed to provide a quality localisation of where corrosion is in the image.

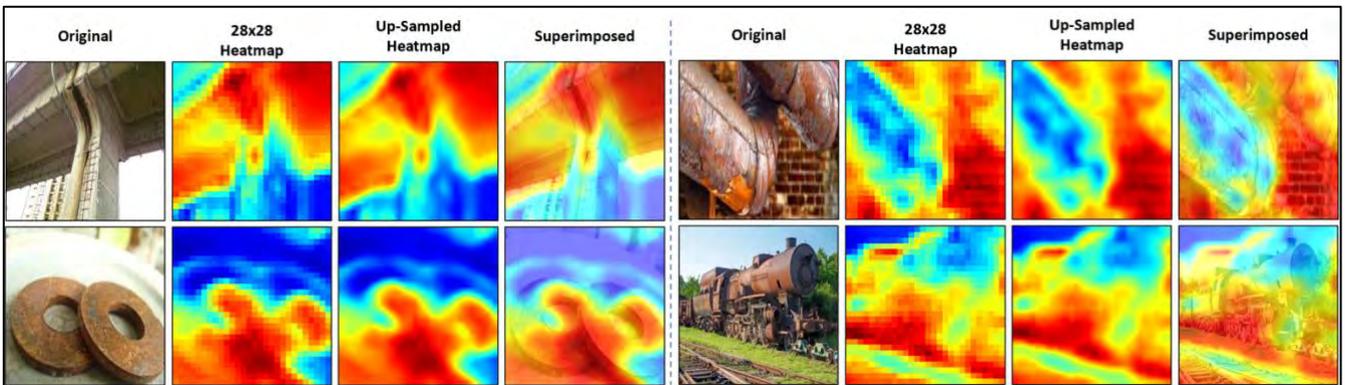
*Figure 10. A random number of samples that reveal relatively poor localisation*

### *Pure CRF Results*

This is the final stage of post processing prior to the results being presented to the user. Again, due to the lack of publicly available per-pixel labelled segmentation datasets for corrosion there is no standard method for benchmarking implementation quantitatively. Instead, analysis was performed qualitatively. The post-processing method of RustSEG is effective at taking coarse heatmaps, and then generating pixel level accurate masks. Examples of this are depicted in Figure 11, which reveals the powerful nature of RustSEG.

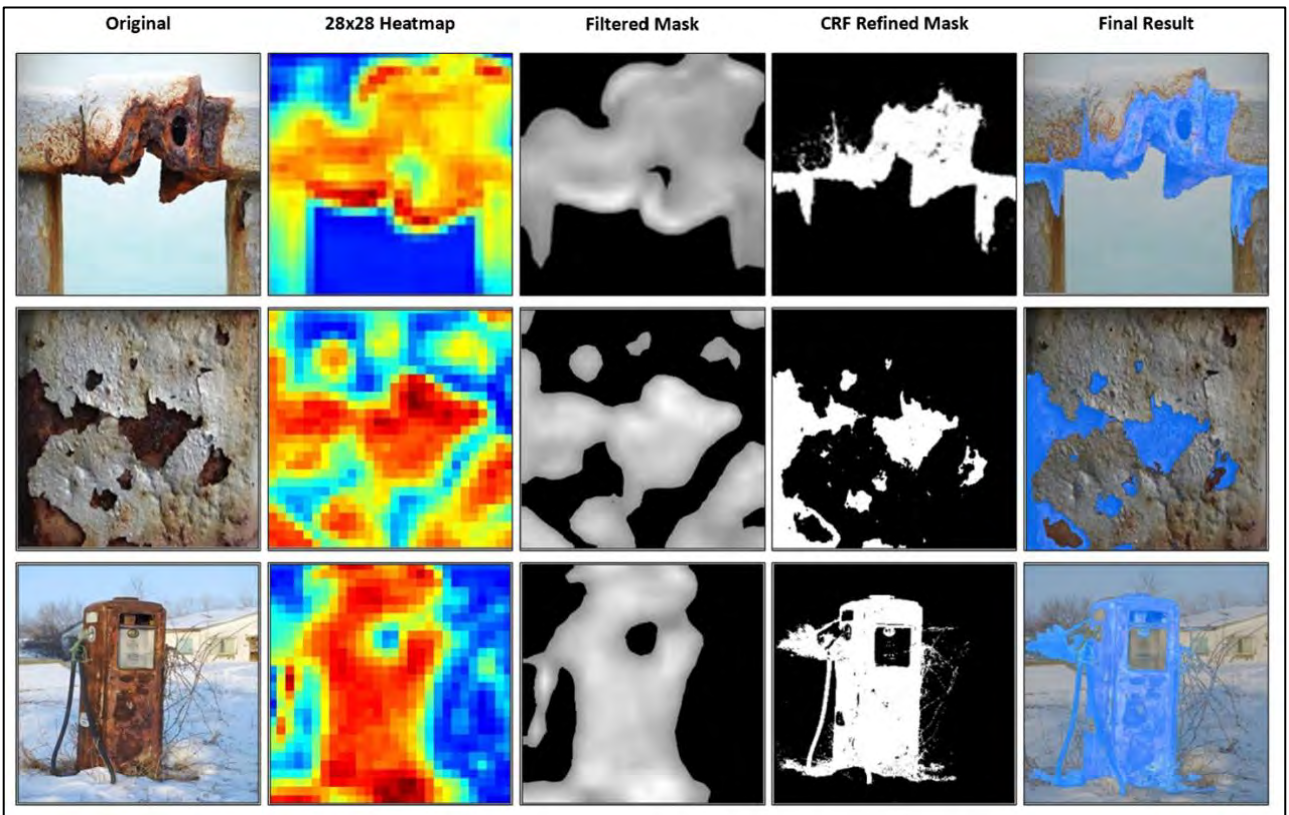

*Figure 11. Good end-to-end results obtained using RustSEG method*

As may be noted from the example of the corroded fuel dispenser (USA gasoline pump), by the nature of the pixel level transformation, the authors determined that the process can be prone to producing larger masks (which would translate to regions of so called 'false positives'). In additional examples where the performance of the accurate mask may be frustrated, *Figure 12* reveals images where the CRF tuning struggled to correctly refine the mask - despite a 'good' initial seed from the heatmap mask. The examples in Figure 12 represents how difficult it is to tune CRFs without learning data to automate the process.

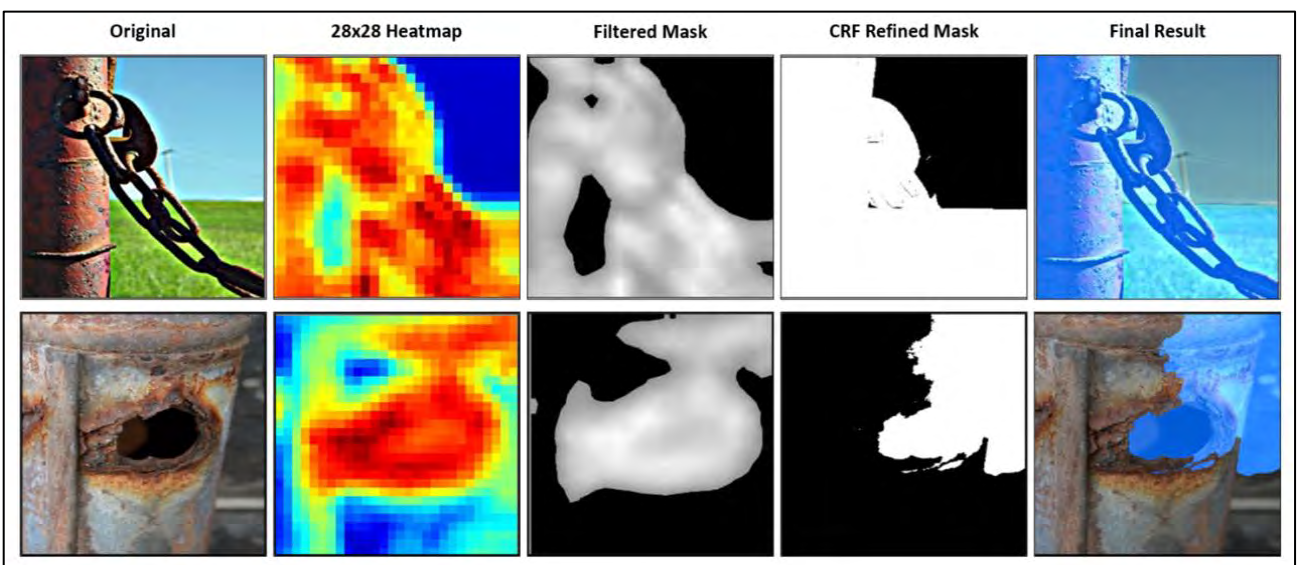

*Figure 12. Examples of post processing method failing to generate accurate masks despite good localisation seeds*

It is noted that post processing can only refine masks that correctly localise corrosion in the initial heatmap stage. The post processing operates on a "junk in – junk out" methodology. In Figure 13 one may see an example with a poor mask seed also returning poor results after final processing by the CRF.

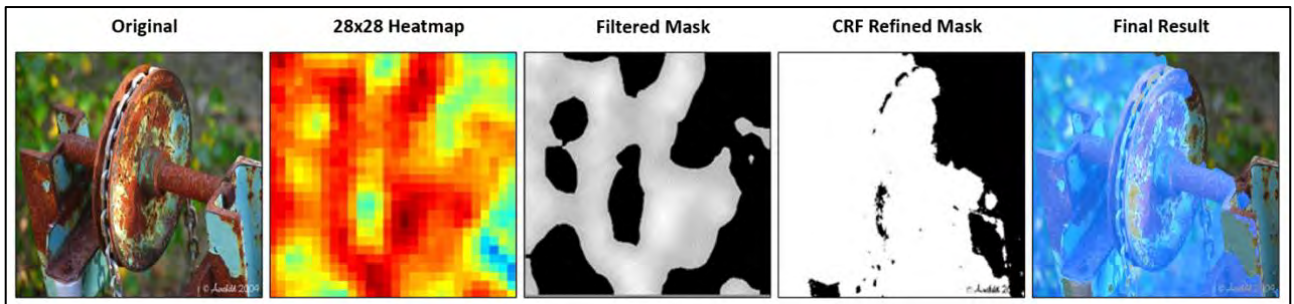

*Figure 13. Example showing that CRF method can not fix poor masks from localisation step*

### *K-means + CRF Results*

Whilst not part of the present instance of RustSEG, the performance of the *K-means + CRF Results* is included in this manuscript, in order to reveal the potential of alternative post processing techniques. This method remains under investigation as a lighter weight version of the original method, for future field or real time deployment – and is applied to the example from Figure 12, as depicted in Figure 14.

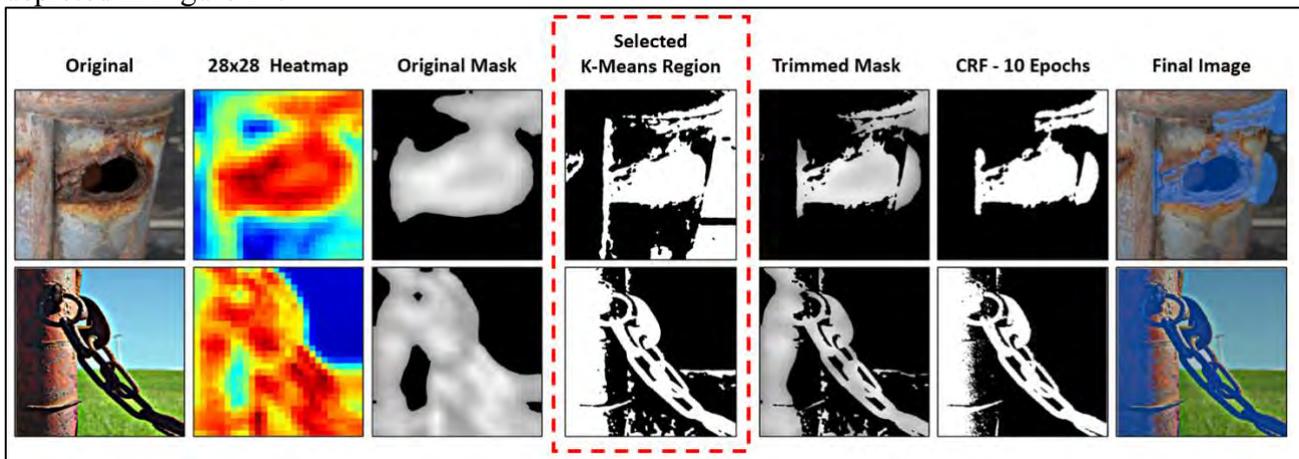

*Figure 14. End-to-end examples where the K-means method outperforms the regular (pure CRF) method.*

The *K-means + CRF Results* method works best in cases where the heatmap generated by the localisation has a true positive region larger than the false positive area. For example, in Figure 15 one can see an example where this assumption causes an erroneous final mask when it does not hold.

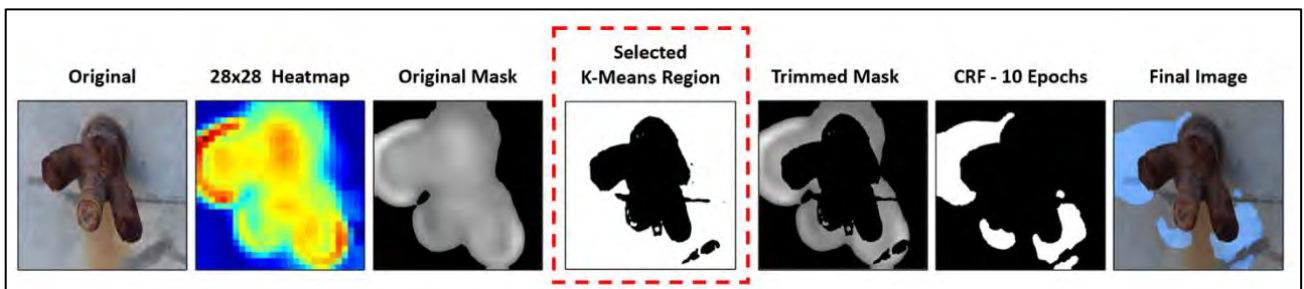

*Figure 15. K-means post processing method where large false positive region triggers wrong K-means mask*

To provide readers with more examples, some additional end-to-end results are provided in the Supplementary Information portion of the paper.

*Inference Times*

Profiling of the overall code and subcomponents of the RustSEG method was undertaken on a range of hardware. Both GPU accelerated hardware, as well as modern multi core CPUs, were examined. The results of the RustSEG model performance are presented in Table 2.

*Table 2. Inference/ run times on different hardware. [Notes: a) End-to-end includes the overheads such as changing data types, however does not include loading in an image; b) The CRF method has no GPU acceleration; c) All tests run on Intel Xeon chips were limited to 1 core and 2 threads; d) All GPU and Xeon test were run in the cloud using the Google Colab service; e) Results are averaged over 10 runs with 10 images, all with detectable corrosion; f) Times are for processing a single image.]*

|  | Device | Task | Time (s) |
|---|---|---|---|
| **CPU** | Intel Core i7-8650U @ 1.90GHz (Turbo @ 4.20GHz) | **Classify** | 0.95 |
|  |  | **Localise** (Grad-CAM++) | 9.89 |
|  |  | **Refine** (CRF: 25 Epochs) | 3.83 |
|  |  | **End-to-End** | 14.96 |
|  | Intel Xeon @ 2.20GHz | **Classify** | 2.62 |
|  |  | **Localise** (Grad-CAM++) | 26.13 |
|  |  | **Refine** (CRF: 25 Epochs) | 2.63 |
|  |  | **End-to-End** | 30.69 |
| **CPU + GPU** | Nvidia Tesla P-100 16GB (PCIe) + Intel Xeon @ 2.20GHz | **Classify** | 0.05 |
|  |  | **Localise** (Grad-CAM++) | 0.18 |
|  |  | **Refine** (CRF: 25 Epochs) | 2.56 |
|  |  | **End-to-End** | 2.83 |
|  | Nvidia Tesla K80 12GB + Intel Xeon @ 2.20GHz | **Classify** | 0.17 |
|  |  | **Localise** (Grad-CAM++) | 0.88 |
|  |  | **Refine** (CRF: 25 Epochs) | 3.16 |
|  |  | **End-to-End** | 4.40 |

The above results indicate this method is not presently implementable in real time – although, it is not far from being able to be. The classification and localisation steps may be parallelized, and this is hence why GPU and multi core CPUs provide better performance. The CRF method cannot be parallelized completely, hence sees little to no performance boost from CPU core counts or GPUs.

**Discussion**

The end-to-end segmentation method herein (RustSEG) can readily produce very accurate segmentation masks of ferrous corrosion on unseen data. This is a key finding, and one that the authors believe is of significant impact. As noted in the preceding results section, several examples were shown of instances and circumstances where the model may not perform to user expectations. Whilst such instances of poor performance are not representative of the typical performance of RustSEG, such instances are relevant to the discussion and how decisions made during the design of the method can affect results.

*Data Set*

*<u>Bias in Training Data</u>*

Although the dataset used for training and evaluation herein is large when compared to other corrosion computer vision methods, it is still small when compared to general deep learning data sets. It is also believed that obtaining the training data from a single source (*Flickr*) created some bias in the training images. As mentioned *flickr.com* is a popular image sharing site for professional photographers, as a result many of images exhibit similar qualities that might have affected the robustness of the training. One example of this is colour and saturation, many of the images in the data set appear to have been edited to make them appear more visually appealing. It should also be noted that using key word searches to find images containing corrosion, creates a bias towards images that contain severe corrosion. It is however, the lived experience of most humans, that instances of corrosion in daily life will be relatively limited compared to most environments in which humans are found. It is also common that for images in which the subject is deliberately corrosion, that the corrosion will be centred and in focus. The latter bias may also go part way to explaining why RustSEG performs so well on images with blurred backgrounds.

The innate bias in the data set was confirmed with the results shown in Table 1. From this data we can see the model performs significantly better when tested on images from *Flickr*, the same source as the RustSEG training set. Additional points in the analysis of the results in Table 1, include the fact that the average quality of images in the *Flickr* test set vs. the expertly labelled data set was significantly larger. Although the images are all scaled to 572×572 before being classified, in cases where the original image is of lower resolution, then the process of scaling up is more likely to distort features like texture. The expertly labelled data set contained significantly more watermarks (~5%). Finally, the expertly labelled data set contains a higher proportion of so-called 'trick' images than the *Flickr* test set, also affecting the comparison of the two.

The process of gaining training data and the biases mentioned herein (involved in obtaining such training data), would suggest that the training dataset is not wholly representative of what 'real world corrosion' may look like. A general corrosion segmentation model is poised to benefit from further high-quality training data.

*<u>The effect of data set size and ratio</u>*

The ratio of corrosion to non-corrosion images was also investigated for this work. It was found for the classifier that a 1:2 ratio performed best for the classifier. However, for localisation of corrosion using Grad-CAM and Grad-CAM++ it was found that one has much richer spatial data using a 1:1 split and that the larger the data set the more precise the localisation becomes. It was decided that a small reduction in classification accuracy would be worth the advantage in localisation so a 1:1 data split was selected. Anecdotal evidence suggests that using one data set ratio for half the training followed by different 'reinforcement' data set with a different ratio could also serve to improve results. Examples of performance with varying data size and ratio are presented in Figure 16.

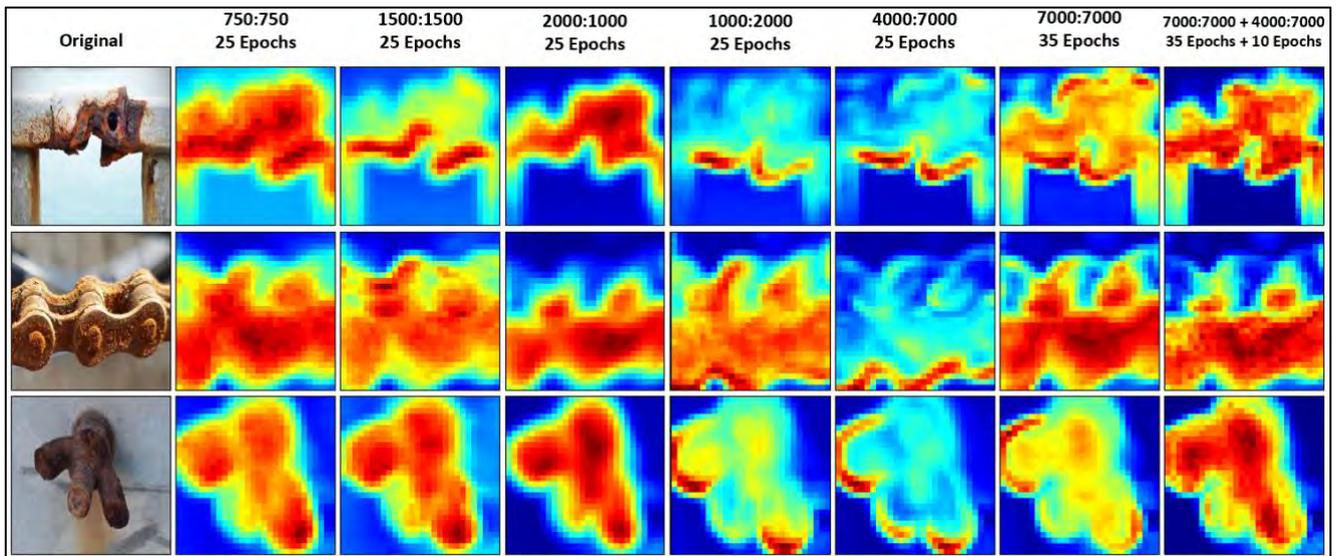

*Figure 16. Small test set example of Grad-CAM++ on the same model with differnt training data set ratios*

### *The Problem of Data Sets*

Although the RustSEG method unequivocally reveals accurate per-pixel segmentation masks that can be generated by only using binary classification datasets – which is to be celebrated – there remains the field wide issue of limited availability and quality of corrosion data sets. If a larger data set, publicly available, of corrosion images was taken from actual (and numerous) corrosion inspections, the RustSEG method would provide better results.

#### *Heat Maps*

The heatmaps generated by the Grad-CAM++ method were deemed as being impressive (even for the authors) for a wide range of input samples. However, in being critically reflective, the method has two drawbacks, the first being computational overhead and the second being resolution.

Initially the RustSEG project had sought to utilise the more widely used Grad-CAM [29] method. A major advantage is that Grad-CAM is more computationally efficient only requiring one pass over the final convolutional layer to calculate the required gradients. However the downsides of Grad-CAM (outlined in [30]) make it inappropriate for the localisation of corrosion. More specifically Grad-CAM has known performance degradation in the analysis of images that contain multiple instances of the same class. For corrosion this was particularly problematic as corrosion is usually not contiguous.

A disadvantage of CAM methods is that they are limited by the size of the final layer of the CNN. To increase to spatial resolution of the localisation method would require increasing the input size of the CNN classifier network. This would not only increase the computational cost of training and classifying images, but it would also significantly increase the computational requirements for the localisation step. Any increase in resolution corresponds to a 3 times increase in the number of gradients required to be calculated.

#### *Post Processing*

From the authors perspective, the post-processing is the most challenging component of the RustSEG method. The robustness of the accuracy of RustSEG is limited by the tuning of the CRF. Not being able to set parameters dynamically or learn parameters means that a "one size fits all" approach was taken. As a result the final post processing mask works very well in some settings and less well in others. In addition, CRFs are computationally expensive.

The K-means approach (briefly presented) demonstrates a practical combination of the heatmaps and the final pixel level segmentation, may be achieved. The CRF makes intuitive sense to be deployed for small improvements to a mask. A method that can take the low detail heatmaps to a medium detail mask for the CRF to take the mask the final step may be a more practical approach in future refinements.

### *The RustSEG Tool and User Website*

From the outset of this work, the principal goal was to create segmentation method that all people could try and use (with appropriate levels of user experience). The purpose of this was twofold. Firstly, and much like the success of corrosiondetecter.com, the hope is that any user facing method can also be used to expand the available corrosion data sets. Secondly, there remains a desire to promote research in this topic area. Corrosion remains a global trillion-dollar per year problem that has major consequences. By making this research publicly available, including access to code and datasets, other researchers and industry can apply methods to improve the RustSEG workflow with the eventual long-term goal of automatic defect detection in infrastructure.

To this end the RustSEG method can be interacted with by users at RustSEG.com (see Figure 17). This mobile friendly website allows users to upload an image (or take one directly from their smartphone camera) and then the RustSEG method will analyse the photo and return its results to the user. Currently the website is hosted using a CPU only server provided by *Amazon Web Service*.

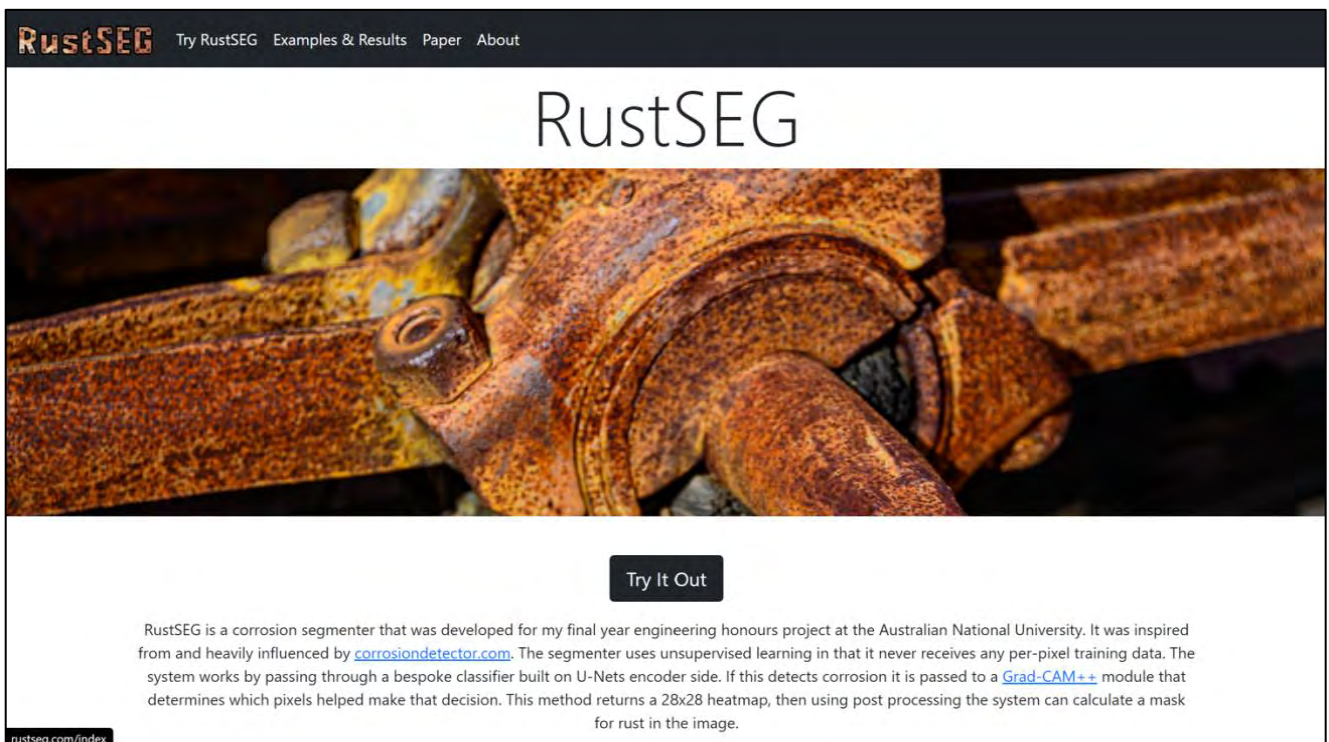

*Figure 17.* *Screen shot of RustSEG.com home page.*

RustSEG.com runs the exact same python implementation of the RustSEG method herein, utilising the Flask API to render the HTML pages. An example of the performance of RustSEG.com is shown in Figure 18, and readers are encouraged to explore the tool and provided feedback.

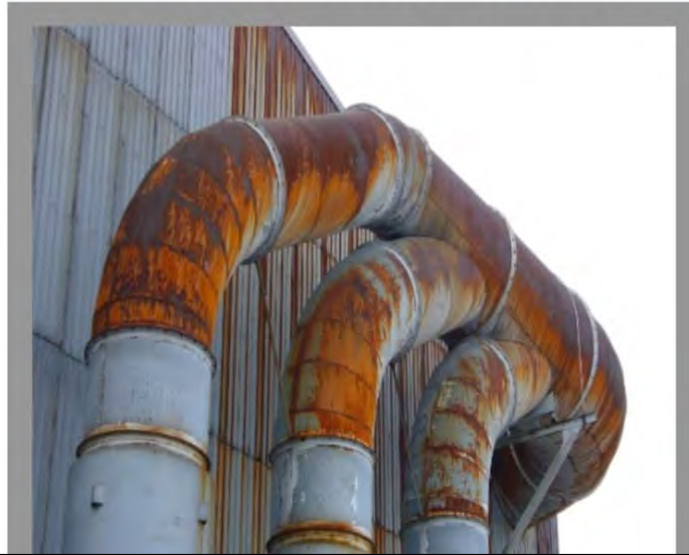
*Figure 18. Screenshot of RustSEG.com after it has processed an image.*

### *Further Work*

Although the results are impressive, the major outcome of this is to serve as a proof of concept. This work shows that it is possible to develop a deep learning based segmentation process for corrosion, using no per-pixel training data. To improve the accuracy of the segmentation masks generated there are many methods could be applied to this workflow that may warrant possible further study in this space.

#### *Improved Diverse Data Set*

A key limiting factor identified for this method was the training dataset used. All the images of corrosion collected were from *Flickr*. However, after separating the data sets into training, validation and test sets only ~6000 corrosion images were available for segmentation. These all ended up being from the single source of *Flickr.com*. As discussed, this data set had bias towards extreme corrosion and ended up making the corrosion classification method less accurate than similar methods like the one described in [17]. Undoubtedly, all deep learning corrosion methods could be improved by curating a more diverse generic data set. This could still include the large number of images publicly available, but could be supplemented with user taken and crowdsourced data, real world inspection images or simply by a researcher building the data set manually (which whilst tedious, remains important).

#### *A Public Segmentation Test Data Set*

Although this work casts doubt on the requirement for cumbersome and expensive per-pixel labelled datasets, the analysis of results does highlight the need for a standard comparison test data set for corrosion segmentation. Standard high performing semantic segmentation methods such as FCN [9] can be compared by applying them to generic segmentation dataset like PASCAL VOC [39], which have publicly available comparison test data sets. This allows researchers to compare methods with like-for-like metrics. This is not presently possible for corrosion segmentation methods. For the simple task of evaluation and comparison this test set would not need to be very large (~50-100 images).

### *Improved Post Processing*

The deep learning based classification and localisation part of the RustSEG workflow performs well, however the final post processing step often includes false positive regions (overestimating corrosion). Also due to (computational) limitations of the Grad-CAM++ method, the original heatmaps have a max region of 28×28 limiting their ability to produce per pixel maps. Additionally, the method of choice for RustSEG, a CRF is computationally heavy, requires complex tuning and doesn't always yield desired results. More accurate methods of post processing the returned heatmaps could be developed. Namely fast methods of classical segmentation could be combined with the initial stages of RustSEG to make the method quicker whilst possibly improving accuracy. It is noted that unless a GPU is available, the limiting factor in speed is not the post processing. From Table 1 it is seen that if a new post processing method may be used, then real-time segmentation (5-10 FPS) may be possible with GPU accelerated hardware

### *Large Classification Data Set + Small Segmentation Data Set*

A further area of research could be to investigate how adding a small amount (20-30 images) of highly accurate per pixel segmented data could augment the method to improve accuracy. For this method two approaches are proposed:

The first would be to retain the existing RustSEG workflow, but use the small labelled data set to automatically tune (train) the CRF post processing stage. For example, in the work the RustSEG's CRF is based upon [42], the authors outline how this improved process may work - and even reveal impressive results on certain data sets.

The other approach would be to add a small segmented data set in the initial CNN training stage. This mixed semi supervised approach has shown promise in other CNN's to boost accuracy with limited data sets[44].

### *Multi-Class Corrosion Classification & Segmentation*

Another area of further investigation for the RustSEG workflow would be to expand the number of classes. In real world corrosion inspections not all corrosion detected is identified as in need of immediate repair. Corrosion is graded into different levels of concern. Currently RustSEG is based on a binary classifier, like the model it is based upon [17]. If corrosion is present, no matter the severity or amount, the binary classifier is triggered. Possible multiclass classifiers could be developed and tested with this workflow to examine if the system is robust enough to highlight corrosion and determine if engineers should be concerned. To investigate this method a multiclass data set separated by severity of corrosion would require to be constructed. Although this would be simpler than per pixel level segmentation, labelling it would still require expert labellers to construct - even after sourcing the images.

**Conclusion**

Corrosion remains a significant issue, both technically and financially. With inspection playing a key role in the management of corrosion, deep learning is an important approach for the development of accurate automatic inspection methods to detect (and then treat corrosion early). This concept extends to various forms of durability (e.g. concrete cracking, coating defects, etc.).

To date, researchers have been unable to produce accurate human level general segmentation methods for corrosion, due to the lack of availability of labelled segmented corrosion data sets. RustSEG - a novel deep learning corrosion segmentation method – was demonstrated, requiring no segmentation training data to generate pixel level segmentation masks with high accuracy. This was demonstrated herein, along with examples and associated discussion.

The proposed RustSEG method (as presented herein) operates by a workflow that unifies existing computer vision methods, employing a "Classify, Localise, Refine" approach. Whilst the performance of RustSEG is not optimised to extend to niche applications or extend beyond human level performance, its performance is considered to be sector leading – and improvements may be made through better training data (inclusive of scenarios akin to those being examined) or modified post-processing techniques. The RustSEG method can be interacted with at RustSEG.com or http://corrosionsegmenter.com/.

**Author contributions**
Ben Burton - conceptualization, methodology, data curation, software, investigation, original draft manuscript preparation, review of manuscript.
Will Nash - conceptualization, methodology, review of manuscript.
Nick Birbilis – conceptualization, data, supervision, preparation and review of final manuscript.

**Supplementary Figures**

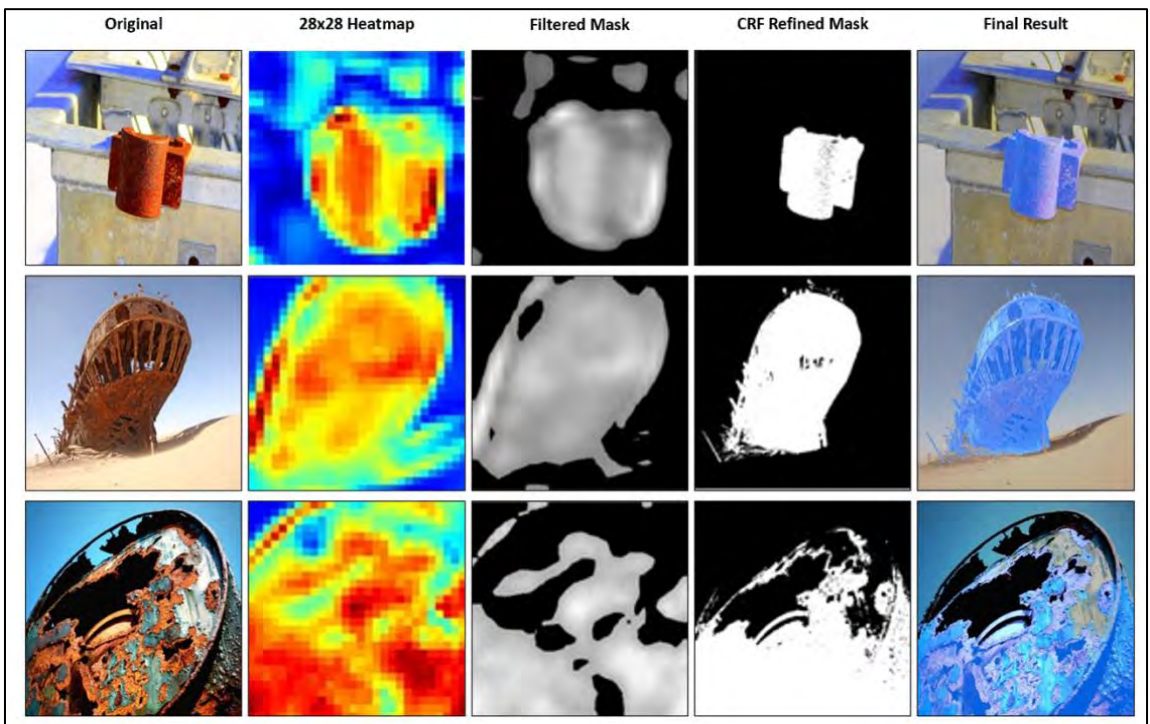

**Figure S1.** Typical end-to-end results for the pure CRF method

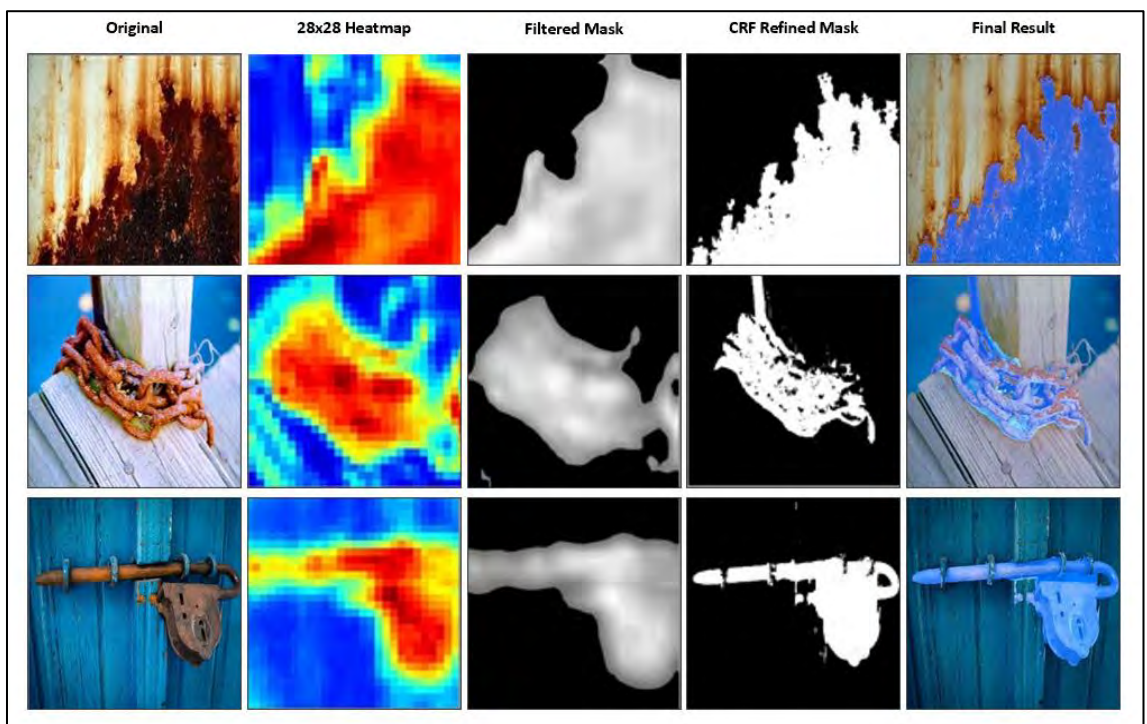

**Figure S2.** Additional end-to-end result examples for the pure CRF method

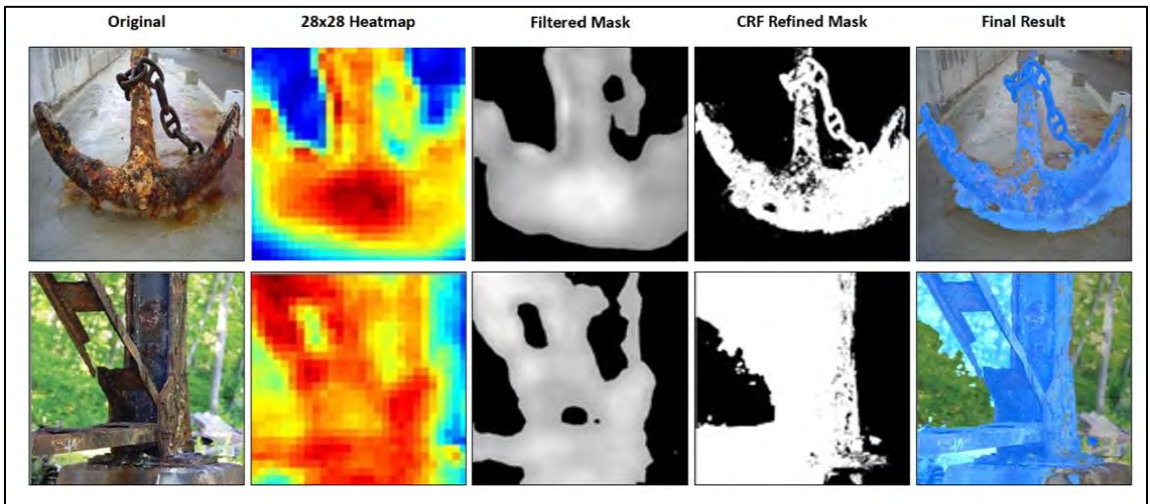

**Figure S3.** End-to-end results that contain evidence of false positive regions (in Final Result)